\documentclass{article}
\usepackage{iclr2027_conference,times}
\usepackage[T1]{fontenc}
\usepackage{amsmath,amssymb,booktabs,array,graphicx,xcolor}
\usepackage{tikz}
\usetikzlibrary{arrows.meta,positioning,calc}
\usepackage{hyperref,url}
\usepackage{cleveref}
\usepackage{graphicx}
\usepackage{subcaption}

\usepackage{booktabs}
\usepackage{colortbl}
\usepackage{xcolor}
\usepackage{siunitx}
\usepackage{fontawesome5}
\usepackage{hyperref}

\definecolor{highlightgray}{gray}{0.92}

\usepackage{booktabs}
\usepackage{multirow}
\usepackage[table]{xcolor} 
\usepackage{siunitx}
\usepackage{graphicx}

\usepackage{wrapfig}

\definecolor{panelgray}{rgb}{0.94, 0.95, 0.96}      
\definecolor{highlightblue}{rgb}{0.93, 0.96, 0.99} 

\hypersetup{colorlinks=true,linkcolor=blue!45!black,citecolor=blue!45!black,urlcolor=blue!45!black}

\definecolor{pendingcolor}{HTML}{955017}

\title{Intuition with Evidence: Adaptive \\ Dual Memory for Multimodal Agents}

\author{%
  \begin{minipage}{0.97\textwidth}
    \centering
    \normalfont
    Qiuhe Hong\textsuperscript{1,}\thanks{Equal contribution.} ,
    Jingya Wang\textsuperscript{1,}\footnotemark[1] ,
    Cunzhen Geng\textsuperscript{2},
    Shenyao Peng\textsuperscript{2},
    Munan Ning\textsuperscript{1}, \\
    Yuyang Liu\textsuperscript{1,2,}\thanks{Corresponding Authors, emails: <\{liuyuyang13, yhtian\}@pku.edu.cn>.} ,
    Yonghong Tian\textsuperscript{1,2,}\footnotemark[2]
    \\[0.8em]
    \textsuperscript{1}School of Electronic and Computer Engineering, Peking University Shenzhen Graduate School\\
    \textsuperscript{2}School of Artificial Intelligence for Science, Peking University Shenzhen Graduate School
  \end{minipage}%
}

\iclrfinaltrue
\begin{document}
\maketitle

\fancyhead{}
\lhead{Preprint}
\renewcommand{\headrulewidth}{0.4pt}

\begingroup
\renewcommand{\thefootnote}{}
\footnotetext{This work was supported by National Natural Science
Foundation of China (62606021).}
\endgroup

\begin{abstract}

As multimodal agents engage in sustained interactions, long-term memory serves two distinct but complementary functions: fine-grained evidence recall and internalized interpretation for global understanding. Although episodic and latent memory address these demands through explicit historical records and internalized experience, neither paradigm alone can accommodate the diverse memory affinities across tasks, owing to limitations in episodic evidence coverage and integration, and in latent factual grounding. To bridge this divide, we propose a dual-memory framework, unifying a model-native latent memory as an intuitive prior with a multi-indexed episodic memory for exact evidence reconstruction via retrieval-conditioned joint alignment. At inference, the framework adaptively composes these memories through a training-free routing mechanism driven by Jensen-Shannon Divergence (JSD) sensitivity and confidence proxies for query-dependent memory affinity. Extensive evaluations across multimodal benchmarks demonstrate state-of-the-art performance, outperforming single-paradigm baselines overall (+4.67\% over vanilla Episodic), with further analysis highlighting episodic strengths in evidence-localized tasks and latent contributions to integrating dispersed information and compensating for imperfect retrieval. Implementation code is avaiable: \href{https://github.com/username/repo}{\faGithub\ lluosi/Intuition\_with\_evidence}.

\end{abstract}

\section{Introduction}
\label{sec:introduction}
As multimodal agents transition to sustained interactions, their long-term memory requirements diversify to encompass highly distinct but complementary objectives including  precise recall of verifiable historical evidence~\citep{wu2024longmemeval} and  internalization of cross-session experiences into  intuitive understanding~\citep{park2023generativeagentsinteractivesimulacra}.
Although effective reasoning relies on both, they impose fundamentally divergent organizational and computational paradigms: exact recall demands recoverable and discrete records to ensure factual fidelity, while contextual reasoning requires abstracting sparse trajectories into a compressed understanding, seamlessly aligned with the model's native comprehension capabilities~\citep{joren2025sufficient}.

Existing single-paradigm architectures fail to satisfy both demands. 
While episodic memory offers interpretable retrieval and extracts semantics via hierarchical reflection~\citep{packer2023memgpt, jiang2026xskill}, explicit retrieval risks incomplete evidence~\citep{wu2024longmemeval} and context dilution~\citep{liu2023lost}. Critically, even given sufficient information, models struggle to integrate disjointed records for organic reasoning~\citep{amiraz2025distracting}.
Conversely, while latent memory internalizes histories into continuous representations for efficient cross-instance transfer~\citep{feng2026elasticmem, zhang2025memgen}, this  compression inherently suffers from information entangles~\citep{zhang2026nextmem} without targeted record patching~\citep{wang2024wise}.

Crucially, real-world interactions exhibit highly diverse \textit{memory affinities}, varying in their demand for specific facts versus global understanding. Thus, no single paradigm suffices for long-horizon tasks.
Recent work coordinates episodic and latent memories mainly by converting explicit records into parametric weights via exclusive routing~\citep{memverse2025, unimem2026}, ignoring the most need for concurrent, synergistic collaboration.
These limitations raise: \textit{Can diverse mechanisms be unified regarding characteristics for varying query-specific memory affinities?
}

\begin{figure}[t]
\centering
\includegraphics[width=1\linewidth]{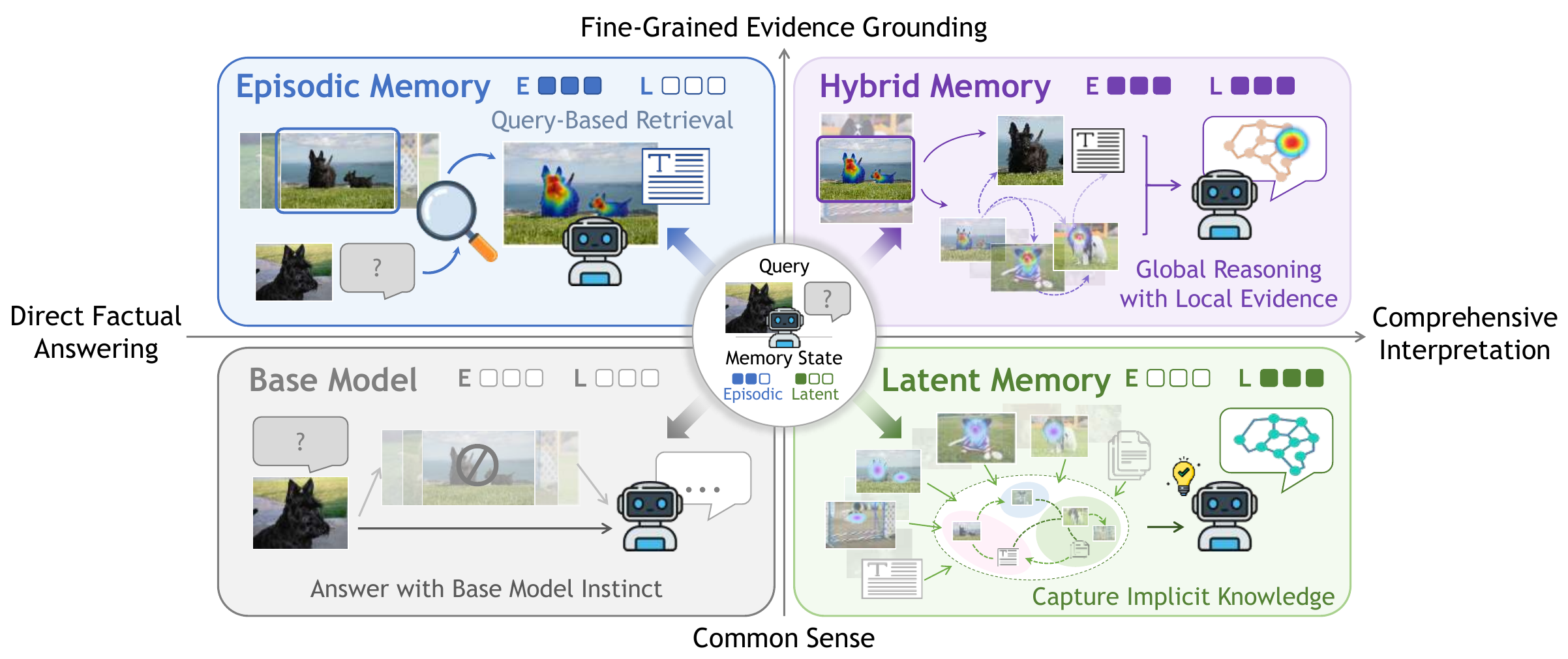}
\caption{
\textbf{Complementary memory paradigms for long-term agent reasoning.} Episodic memory reconstructs source-grounded evidence for factual recall, while Latent memory internalizes trajectories into into model-native representations for global reasoning.
As neither is universally dominant, our framework maintains both mechanisms and adaptively composes them at inference.
}

\label{fig:motivation}
\end{figure}

We propose a dual-memory framework that synergizes \textit{cognitive intuition} with \textit{factual verification} by decoupling underlying memory roles.
Specifically, a multi-indexed episodic store retrieves multimodal observations for precise \textit{evidence grounding}, whereas an inference-side native latent memory internalizes trajectories via test-time adaptation to establish an \textit{intuitive prior}.
Rather than operating as isolated components, they are synchronized through joint alignment training for cohesive reasoning.
During inference, further to dynamically match query-specific memory affinities conditioned on the current memory state, we deploy a training-free adaptive router driven by JSD sensitivity and sequence confidence proxies, unifying these mechanics into an organic memory framework.


We evaluate our framework on multi-turn multimodal benchmarks across various model scales to validate the effectiveness of hybrid memory orchestration.
Beyond notable inference efficiency and robustness, our analysis uncovers distinct task-specific memory affinities, underscoring the necessity of co-utilizing complementary memory mechanisms organically.
Our contributions are:
\begin{itemize}
    \item We systematically analyze the inherent failure modes of single-paradigm memories (episodic vs. latent vs. fixed hybrid) and uncover distinct task-specific memory affinities.
    \item We propose a novel dual-level memory framework combining episodic evidence grounding with native latent intuitive priors, adaptively orchestrated by  adaptive proxy-guided routers.
    \item We conduct extensive experiments to validate the effectiveness, robustness, and efficiency of the hybrid memory mechanism with in-depth analysis for memory characteristics.
\end{itemize}

\section{Related Work}
\label{sec:related_work}

\paragraph{Episodic Memory.}
LLM agents commonly use external textual memory by incorporating retrieved records into the input context~\citep{park2023generativeagentsinteractivesimulacra}.
To support efficient access, these records are organized into structured stores, including hierarchies and graphs, with retrieval performed before generation or interleaved with reasoning~\citep{chhikara2025mem0,xu2025amem,yan2025memoryr1}.
Multimodal extensions combine textual abstractions with visual representations, supporting retrieval of both semantic information and fine-grained perceptual evidence~\citep{long2025seeinglisteningrememberingreasoning,kang2026vmem}.
However, query-driven retrieval can miss implicitly relevant memories, while broader retrieval increases inference costs~\citep{hu2026evermembench,hu2025memoryagentbench}.
Even accurate retrieval may leave a semantic gap between human-readable representations and the model's in-context interpretation and reasoning~\citep{wu2024longmemeval,shen2026mem2actbench}.

\begin{figure}[t]
    \centering
    \includegraphics[width=1\linewidth]{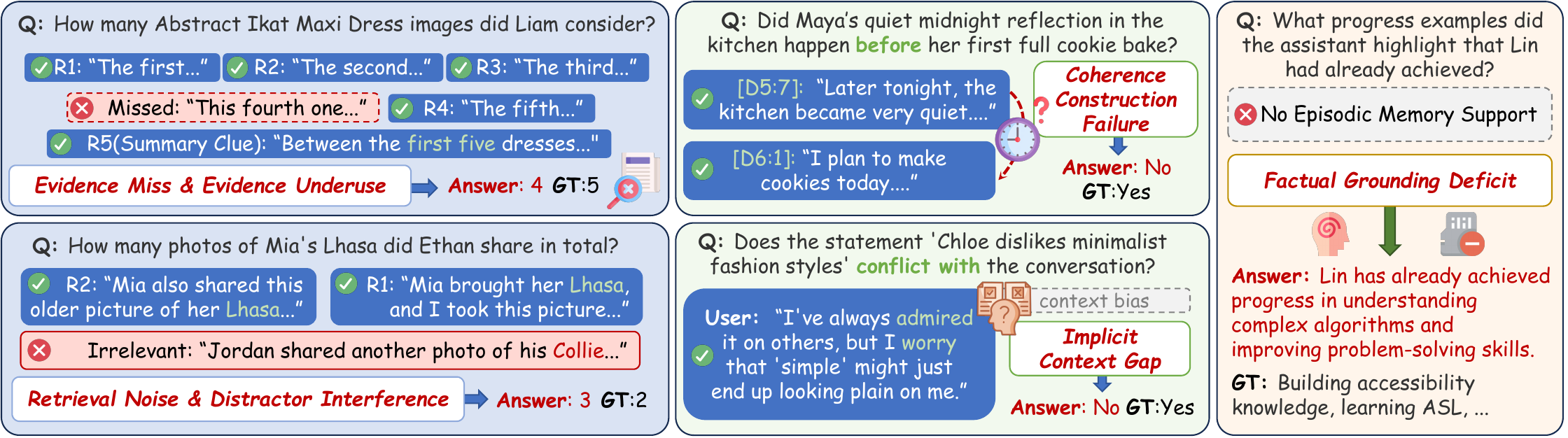}
    \caption{\textbf{Failure cases of individual memory branches conducted on Mem-Gallery~\citep{memgallery2026}.}
\textit{Left}: Episodic memory is limited by retrieval quality (omissions and distractions).
\textit{Middle}: Episodic memory is limited by contextual understanding (evidence integration and implicit alignment).
\textit{Right}: Latent memory fails to recover the specific details required by the query.}
\label{fig:limitation}
\end{figure}

\paragraph{Latent Memory.}

Latent memory encodes experience in model-native representations rather than explicit records.
Some retrieve cached continuous representations or compress retrieved experience into soft tokens for inference~\citep{feng2026elasticmem, fu2026latent}, while retaining dependence on retrieval coverage.
Other internalize experience through training, learning either context-dependent memory generators~\citep{zhang2025memgen} or reusable procedural and tool knowledge~\citep{wu2026tokmem,wang2025toolgen} without light adaptation to new scenarios.
Multimodal variants generate latent perceptual and semantic memories~\citep{yu2025vismem}, or verbalize parametric recall, retaining dependence on in-context reasoning~\citep{liu2025memverse}.
However, preserving broad semantic content does not ensure factual fidelity as distorting details under capacity constraints~\citep{zhang2026nextmem}.

\paragraph{Test-Time Adaptation.}
Test-time learning enables models and agents to adapt to incoming data and feedback during deployment.
Early work studies self-supervised parameter updates~\citep{sun2020ttt}, with recent extensions to LLMs~\citep{feng2026inplace}.
It further supports context retention through recurrent and hierarchical neural memories~\citep{behrouz_titans,behrouz2025hope}, as well as compression into adapters~\citep{chen2025perk}.
For agents, test-time adaptation can also occur at the harness level by accumulating reflections and revising external memory without weights drifting~\citep{shinn2023reflexion,xu2025amem}.
These developments motivate coordinated adaptation across parametric and external memory.
UniMem consolidates recurring episodic experience into parametric procedural memory, but its mutually exclusive routing ignores the characteristics between two memory~\citep{unimem2026}.
Building on these works, we study how combining the complementary dual-level memories can improve test-time adaptation.

\section{Rethinking Long-Term Memory: Evidence or Intuition}
\label{sec:dual_memory}

\subsection{Motivation: Uni Memory Failures in the MEMORY AFFINITY GAP}
\label{sec:dual_rationale}

Episodic and latent memory, two major memory paradigms,  operate through distinct mechanisms and have their own  limitations, as illustrated by the case studies in \Cref{fig:limitation}.

Episodic memory can fail in both evidence coverage and interpretation.
On the retrieval side, beyond missing direct evidence, query similarity matching can omit supporting cross-episode background and relations~\citep{gutierrez2024hipporag}.
Broadening retrieval without adequate context may introduce distraction~\citep{amiraz2025distracting}.
On the inference side, even with sufficient evidence, 
the model may fail to organize and combine it correctly~\citep{joren2025sufficient}.
Interpretation may further require drawing on implicit global semantics and stylistic conventions distributed across the history.

Latent memory internalizes experience within a limited parameter budget. Semantic abstraction may blur distinctions between episodes, hindering recall of the specific entities and events relevant to a query.
Even when the relevant experience is recalled, the model may reinterpret and stylistically reformulate its content, compromising recovering to the original evidence.
These failures suggest proper combination of distinct memory for complementary.

\subsection{Method: Episodic Memory as an Evidence Layer}
\label{sec:external_layer}
Episodic memory $\mathcal E$ supports inference with query-relevant
textual and visual evidence.
We organize past trajectories into episodes, each containing
original records $r\in\mathcal R$ augmented with derived 
memories $\mathcal M$ at multiple granularity and across modalities that serve as retrieval indices:
\begin{equation}
     \mathcal E=(\mathcal R,\mathcal M),
    \qquad
    \mathcal M
    =\{\mathcal M_i\}_{i\in
    \{\mathrm{sum},\mathrm{evt},\mathrm{sem},\mathrm{res}\}}.
    \label{eq:episodic_store}
\end{equation}
Summary memory $\mathcal M_{\mathrm{sum}}$ groups summaries,
topics, and entities by chunk.
Event and Semantic memories $\mathcal M_{\mathrm{evt}}$ and
$\mathcal M_{\mathrm{sem}}$ organize event summaries and
cross-episode facts and preferences, respectively, into
semantic clusters. Resource memory $\mathcal M_{\mathrm{res}}$ maintains a flat collection of images.
Source links connect these derived memories to their
supporting original records.

Given a query $q$ containing text and an optional image,
a modality router selects text-only or multimodal retrieval.
Text retrieval combines matching within
episodes identified via $\mathcal M_{\mathrm{sum}}$ with
global searches over $\mathcal M_{\mathrm{sem}}$ and
$\mathcal M_{\mathrm{evt}}$.
When required, image retrieval matches $q$ against images in $\mathcal M_{\mathrm{res}}$. Matched $\mathcal{M}$ contribute relevance scores $s$
to the corresponding $r$. Each $r$ then receives
a combined score $S(q,r)$ across retrieval channels:
\begin{equation}
\begin{aligned}
S(q,r) ={}&
\lambda_{\mathrm{sum}}s_{\mathrm{sum}}(q,r)
+\lambda_{\mathrm{evt}}s_{\mathrm{evt}}(q,r)+\lambda_{\mathrm{sem}}s_{\mathrm{sem}}(q,r)
+\lambda_{\mathrm{res}}s_{\mathrm{res}}(q,r),
\end{aligned}
\label{eq:external_retrieval}
\end{equation}
where $\lambda$ weights each channel.
The top-$K$ $r$ ranked by $S(q,r)$ constitute the evidence
set $\mathcal R_q$.
For visual retrieval, semantic captions are selectively supplemented with original images, adaptively allocating
additional budget to fine-grained visual evidence.

\begin{figure}[t]
    \centering
    \includegraphics[width=1\linewidth]{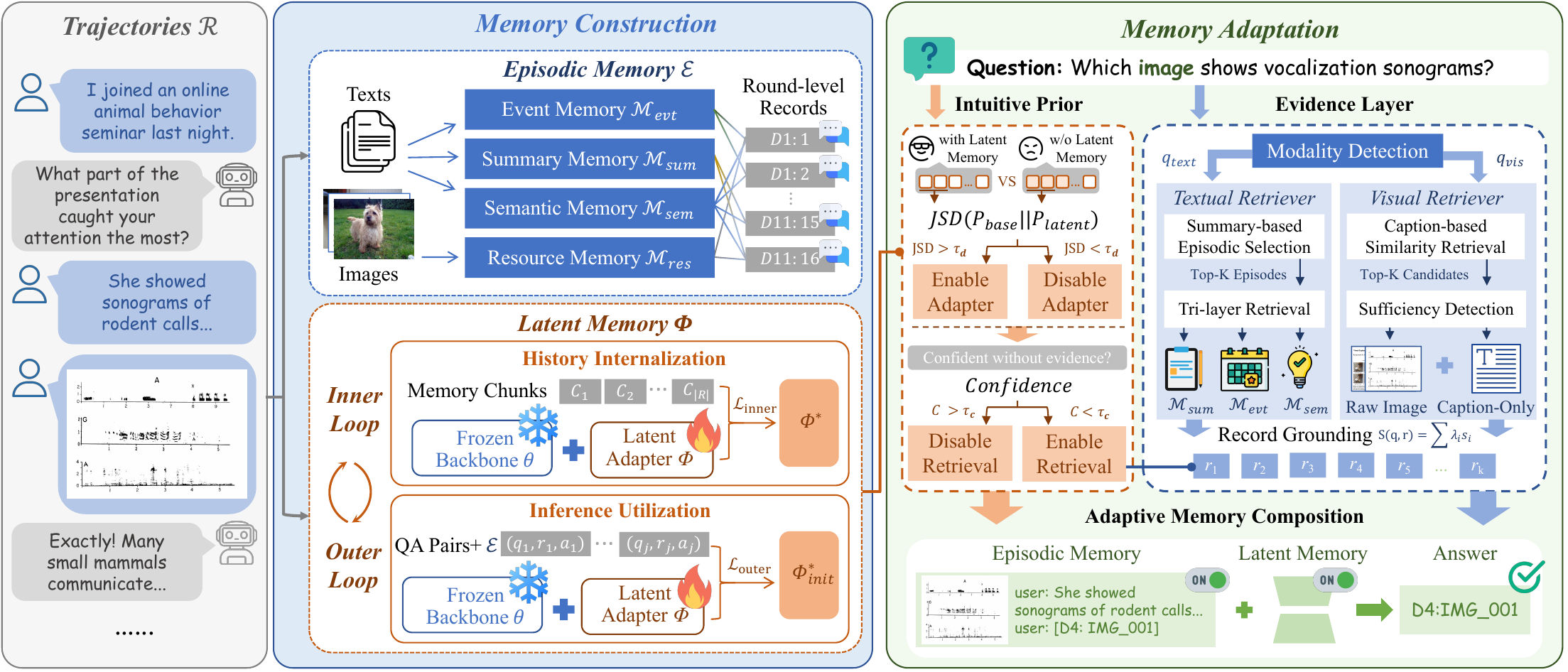}
    \caption{\textbf{Overall Workflow of our method.} Memory construction transforms multimodal trajectories into complementary episodic and latent memories. Given a query, memory adaptation selectively activates and integrates these memories by training-free routers to generate the answer.
    }
    
\label{fig:workflow}
\end{figure}

\subsection{Method: Latent Memory as an Intuitive Prior}
\label{sec:latent_layer}

Reusable, high-level understanding of the past can be internalized in model parameters as latent memory, providing intuitive guidance.
Latent memory is encoded in a trainable adapter $\phi$
attached to the frozen inference backbone $\theta$.
A meta-learning framework is adopted to enable rapid history encoding through test-time training~\citep{sun2020ttt,chen2025perk}.

First, $\mathcal R$ is internalized in $\phi$ via an inner loop that minimizes
a causal language-modeling objective:  
\begin{equation}
    \mathcal{L}_{\mathrm{inner}}(\phi,\mathcal R)
    = \frac{1}{|\mathcal R|}\sum_{r=1}^{|\mathcal R|}
      \mathcal{L}_{\mathrm{NTP}}(\phi; C_r),
    \label{eq:history_encoding}
\end{equation}
where $\{C_r\}_{r=1}^{|\mathcal R|}$ are chunks of $\mathcal R$, with images represented by captions as semantic records, and $\mathcal{L}_{\mathrm{NTP}}$ denotes the next-token prediction loss.
After the inner-loop adaptation, the resulting state $\phi_t'$ is evaluated on a small QA support set \(\mathcal D_{\mathrm{align}}\) to strengthen its use for downstream inference.
For each $(q_i,a_i)\in\mathcal D_{\mathrm{align}}$,
the answer loss is computed with 
$\mathcal R_{q_i}$ retrieved from $E$ to optimize joint memory use:
\begin{equation}
\mathcal{L}_{\mathrm{outer}}(\phi)
=
   -\frac{1}{|\mathcal D_{\mathrm{align}}|}
  \sum_{(q_i,a_i)\in\mathcal D_{\mathrm{align}}}
  \frac{1}{|a_{i}|}
  \sum_{j}^{|a_i|}
  \log p_{\phi}(a_{i,j}\mid q_i,\mathcal R_{q_i},a_{i,<j}).
\end{equation}
The corresponding gradient $\nabla_{\phi}\mathcal L_{\mathrm{outer}}(\phi)$ is then used to update the initialization $\phi$ through a first-order approximation to improve inner-loop encoding:
\begin{equation}
    \phi_{t+1}
    = \operatorname{AdamW_{\eta}}
    \left(
        \phi_t,\,
        \operatorname{ClipNorm}
        \left(
            \operatorname{stopgrad}
            \left[
                \left.
                \nabla_{\phi}\mathcal L_{\mathrm{outer}}(\phi)
                \right|_{\phi=\phi'_t}
            \right],c
        \right)
    \right),
\end{equation}
where $\phi_t$ denotes the initialization at iteration $t$ and $\phi'_t$ encoded state after inner-loop adaptation. Here, $\eta$ and $c$ denote the outer-loop learning rate and gradient clipping threshold, respectively.

\section{Adaptive Memory Composition during inference}
\label{sec:composition}

\subsection{Diverse Memory Suitability}
\label{sec:mode_utility}

\begin{figure}[h]
    \centering

    \begin{minipage}[t]{0.68\textwidth}
        \vspace{0pt}
        \centering

        \includegraphics[
            width=\linewidth
        ]{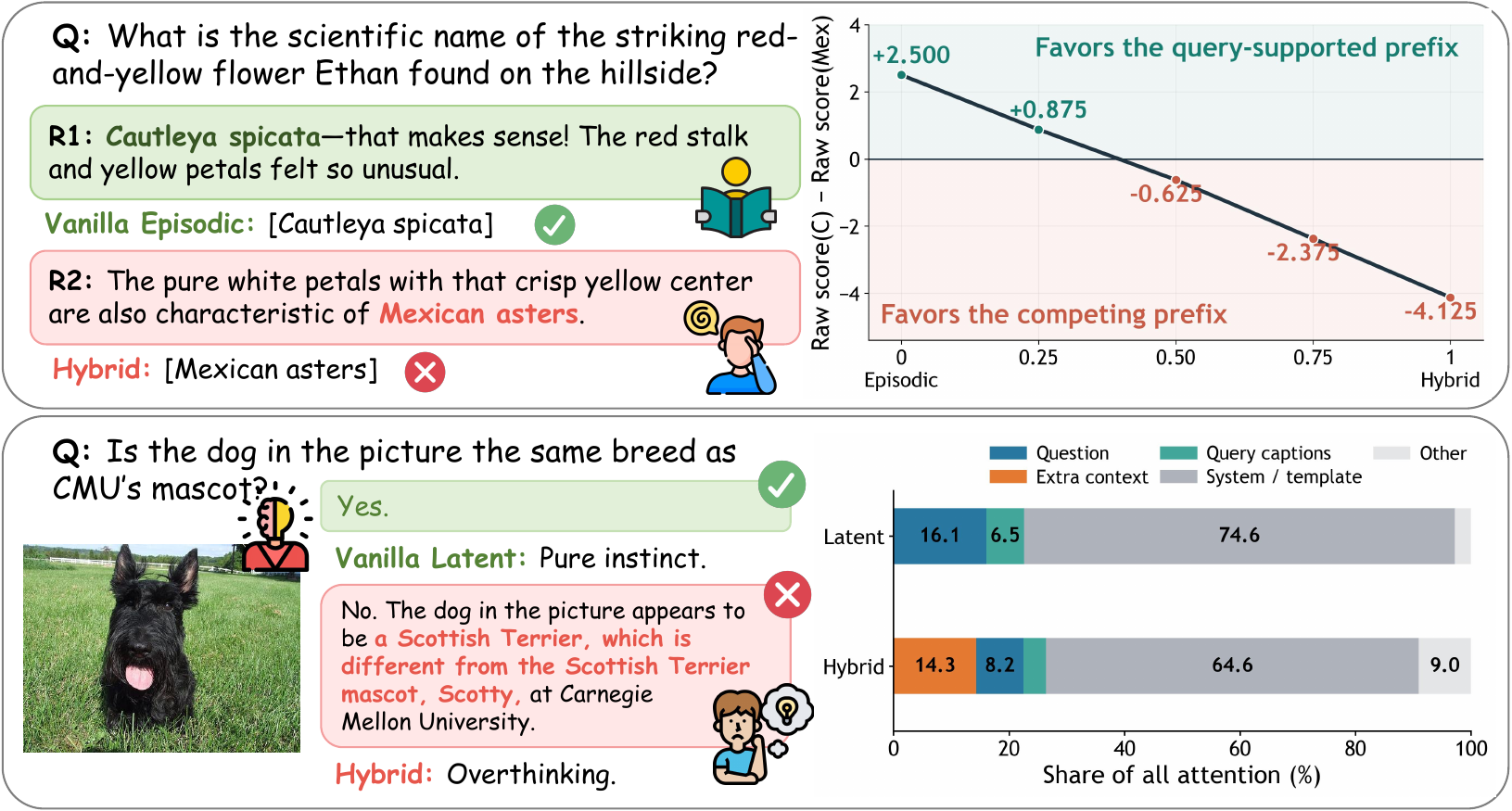}
        \caption{
        \textbf{Uniform hybrid configurations can degrade performance.} 
        \textit{Top}: Increasing hybrid strength draws episodic-locked evidence toward misleading context. \textit{Bottom}: Extra evidence attention load disrupts coherent latent reasoning, leading to self-contradictions.
        }
        \label{fig:left-figure}
    \end{minipage}
    \hfill
    \begin{minipage}[t]{0.30\textwidth}
        \vspace{0pt}
        \centering
        \includegraphics[
            width=\linewidth
        ]{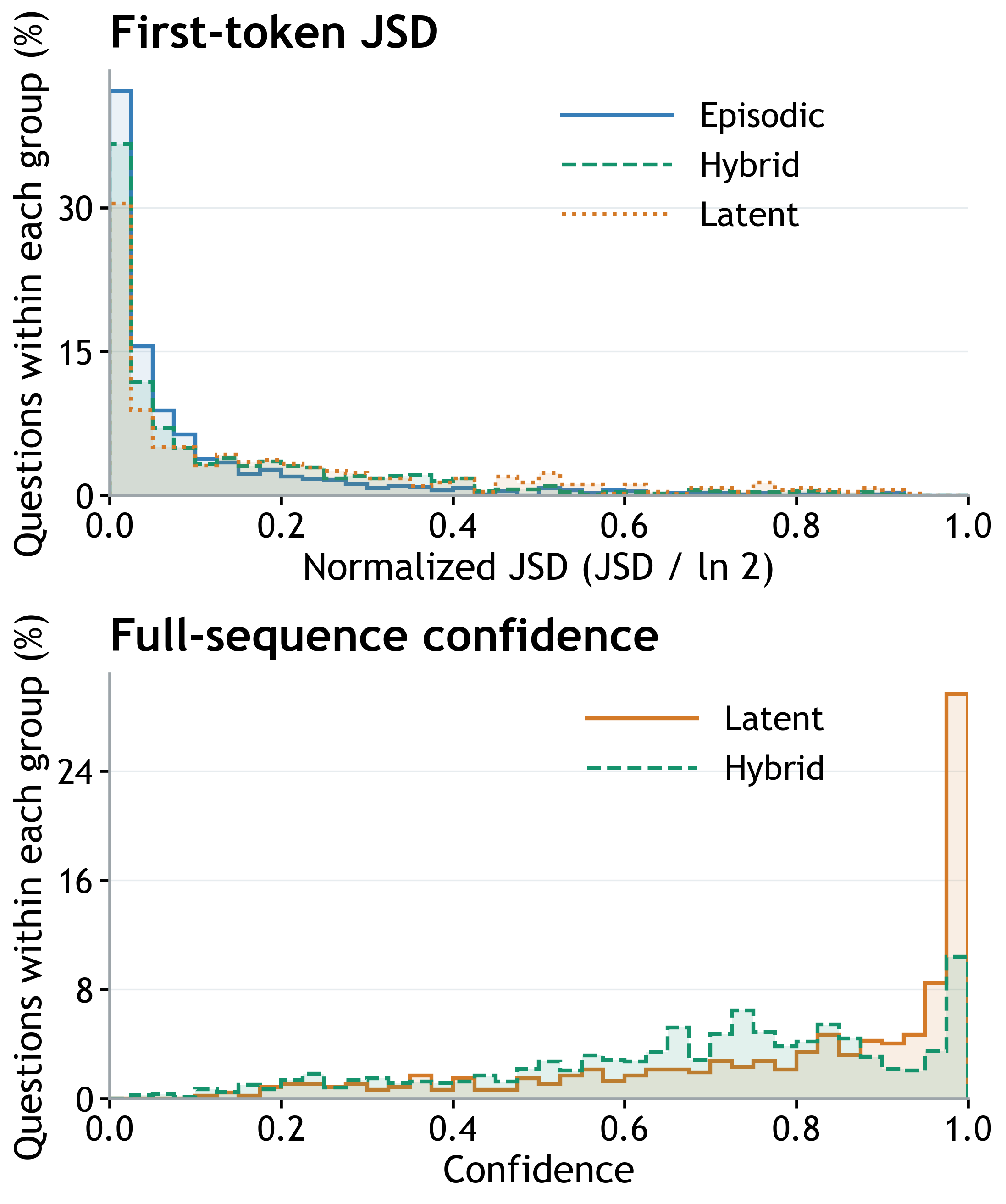}
        \caption{Distributions of first-token JSD and full-sequence confidence across winning samples for each memory paradigm.}
        \label{fig:proxy-distributions}
    \end{minipage}
\end{figure}

Although hybrid memory integrates the complementary properties of episodic and latent representations, different queries may favor different memory mechanisms, motivating adaptive modulation.
As illustrated in \Cref{fig:left-figure}, uniform hybrid inference neglects query–memory compatibility and even can degrade answer quality relative to vanilla inference.
When latent memory already captures the relevant information across past interactions, episodic retrieval may bias the inference toward instances integration at the expense of global coherence.
Conversely, for queries answerable directly from episodic evidence, misaligned latent priors may lead the model to reinterpret the evidence to fit previously learned patterns.
These findings suggest a reliable router for memory composition.

\subsection{Router Proxy}
\label{sec:router_proxy}

To efficiently adapt memory composition to the current query and memory state, a training-free, two-stage proxy mechanism is proposed. 
To assess how well internalized patterns align with the query, the first stage measures sensitivity to latent activation $d(q,\phi) $ by Jensen--Shannon divergence between the base and latent-only
first-token distributions:
\begin{equation}
d(q,\phi) = JSD(P_{\theta}^{(1)}(q) || P_{\theta,\phi}^{(1)}(q)).
\label{eq:latent_responsiveness}
\end{equation}
For queries retaining latent memory, the second stage further evaluates whether internalized knowledge
is sufficient to answer the query using confidence $c_L(q,\phi)$:
\begin{equation}
c_L(q,\phi)
=
\exp\left(
\frac{1}{T}
\sum_{t=1}^{T}
\log p_{\theta,\phi}
\left(\hat{a}_t \mid q,\hat{a}_{<t}\right)
\right).
\label{eq:latent_confidence}
\end{equation}
\Cref{fig:proxy-distributions} compares the distributions of both
proxies across queries grouped by successful memory modes.
Although substantial overlap remains owing to the inclusion of co-winners, episodic-memory successes are more concentrated in the low-JSD region, supporting that vanilla episodic inference is more robust when latent responsiveness is weak.
For confidence, latent winners exhibit a right-shifted distribution relative to hybrid winners with higher confidence for the current query.

\subsection{Overall Framework}
\label{sec:composition_framework}

As \Cref{fig:workflow}, given $\mathcal{R}$, construct $\mathcal E$ and $\phi$  following \Cref{sec:external_layer} and \Cref{sec:latent_layer} respectively.
For a query $q$, we first compute $d(q,\phi)$ as \cref{eq:latent_responsiveness} to determine whether to activate $\phi$ with calibrated $\tau_d$:
\begin{equation}
g_L(q,\phi) =  \mathbb{I}[d(q,\phi) \ge \tau_d].
\label{eq:latent_gate}
\end{equation}
We further compute $c_L(q,\phi)$ via \cref{eq:latent_confidence}
only when $g_L=1$, and define the episodic gate
using $\tau_c$:
\begin{equation}
g_\mathcal E(q,\phi) =  \mathbb{I}[d(q,\phi) < \tau_d \lor c_L(q,\phi)<\tau_c].
\label{eq:epi_gate}
\end{equation}
When $g_\mathcal E(q,\phi)=1$, we retrieve $\mathcal R_q$ according to
$S(q,r)$ in \cref{eq:external_retrieval} and concatenate it with $q$.
Final inference distribution is computed as follows, with
$\mathcal R_q^{(0)}=\varnothing$ and
$\mathcal R_q^{(1)}=\mathcal R_q$:
\begin{equation}
p(a\mid q)
=
p_{\theta,\,g_L(q,\phi) \cdot \phi}
\left(a\mid q,\mathcal R_q^{(g_\mathcal E(q,\phi))}\right).
\label{eq:gated_inference}
\end{equation}

\section{Experiments}
\label{sec:experiments}

\subsection{Setups}

\paragraph{Baselines and Benchmarks.}
We compare our approach against ten representative memory baselines, together with a no-memory setting: (a) \textbf{no external memory}:
Vanilla Base, which performs direct inference with the backbone
model; (b) \textbf{text-only retrieval-augmented methods}: Full Memory(Text), First-in-first-out(FIFO), NaiveRAG, Mem0 \citep{chhikara2025mem0}, and MemGPT \citep{packer2023memgpt}; (c)
\textbf{multimodal retrieval-augmented memory}: Full Memory(Multimodal), MuRAG \citep{chen2022murag}, and UniversalRAG \citep{universalrag2025}; and (d) \textbf{latent
memory}: MemVerse \citep{memverse2025}, which combines long-term graph storage with a trained memory generator, while most alternatives require extensive training data.
Evaluations are conducted on \textbf{Mem-Gallery}
\citep{memgallery2026} and the \textbf{LoCoMo}~\citep{maharana2024locomo} subsets of
MemoryBench~\citep{memorybench2025}, both targeting memory management within multi-turn multimodal conversations, simultaneously emphasizing fine-grained evidence recall and comprehensive reasoning.


\paragraph{Metrics.}
To evaluate memory performance, we employ both lexical and semantic metrics. Specifically, Token-level F1, BLEU-1, and Exact Match (EM) assess lexical answer quality \citep{rajpurkar-etal-2016-squad,papineni-etal-2002-bleu}. For high-level semantic evaluation, we use DeepSeek-V4-Flash~\citep{deepseekai2026deepseekv4} as an LLM judge to score the correctness following Mem-Gallery protocols~\citep{memgallery2026}. and the LLM-as-a-Judge paradigm \citep{zheng2023judging}. 


\begin{table*}[t]
\centering
\footnotesize 
\setlength{\tabcolsep}{8pt} 
\renewcommand{\arraystretch}{1.2} 

\caption{\textbf{Main results on Mem-Gallery and LoCoMo.} Comprehensive evaluation of memory frameworks using Qwen2.5-VL (3B and 7B) across F1, BLEU-1, EM, and LLM-Judge. In addition to text-only and multimodal baselines, we report the Vanilla Base (no memory), standalone Episodic and Latent modules, fixed Hybrid composition, and Adaptive routing.}
\label{tab:main_results}

\sisetup{
  table-format=2.2,
  table-number-alignment=center,
  table-auto-round=false,
  mode=text,
  detect-all
}

\begin{tabular}{l *{3}{S} S[table-format=2.2] | *{3}{S} S[table-format=2.2]}
\toprule
\multirow{2.5}{*}{\textbf{Method}} 
& \multicolumn{4}{c}{\textbf{Mem-Gallery}} 
& \multicolumn{4}{c}{\textbf{LoCoMo}} \\
\cmidrule(r){2-5} \cmidrule(l){6-9}
& {\textbf{F1}} 
& {\textbf{BLEU-1}} 
& {\textbf{EM}} 
& {\textbf{Judge}} 
& {\textbf{F1}} 
& {\textbf{BLEU-1}} 
& {\textbf{EM}} 
& {\textbf{Judge}} \\
\midrule

\rowcolor{panelgray}
\multicolumn{9}{c}{\textbf{Qwen-2.5-VL-3B}} \\
\addlinespace[2pt]

Vanilla Base
    & 29.69 & 26.03 & 18.21 & 36.14
    & 29.36 & 26.16 & 10.00 & 34.00 \\

FIFO
    & 31.12 & 27.83 & 19.78 & 33.53
    & 36.83 & 32.22 & 10.00 & 42.00 \\

NaiveRAG
    & 58.99 & 54.31 & 40.11 & 63.17
    & 25.01 & 20.49 &  6.00 & 37.00 \\

Mem0
    & 29.89 & 26.67 & 19.96 & 31.98
    & 36.94 & 33.02 & 10.00 & 46.00 \\

Full-Context
    & 39.05 & 35.84 & 26.16 & 40.17
    & 35.01 & 31.55 & 16.00 & 50.00 \\

MemGPT
    & 39.06 & 35.84 & 26.16 & 41.13
    & 37.01 & 33.55 & 18.00 & 43.00 \\

\cmidrule(r){1-5} \cmidrule(l){6-9}

Full-MM-Context
    & 33.80 & 30.81 & 22.64 & 34.19
    & 37.37 & 33.51 & 16.00 & 41.00 \\

MuRAG
    & 63.71 & 59.10
    & 44.92 & 67.58
    & 31.41 & 25.93 & 14.00 & 35.50 \\

UniversalRAG
    & 62.95 & 57.90 & 43.81 & 65.75
    & 31.41 & 25.93 & 14.00 & 35.50 \\

MemVerse
    & 32.30
    & 28.96
    & 20.69
    & 37.48
    & 20.88
    & 16.06
    & 6.00
    & 42.50 \\

\cmidrule(r){1-5} \cmidrule(l){6-9}

\rowcolor{highlightblue}
Episodic
    & 62.90 & 57.96 & 43.16 & 67.28
    & 38.30 & 32.30
    & 16.00 & 42.00 \\

\rowcolor{highlightblue}
Latent
    & 38.87 & 34.95 & 25.69 & 38.54
    & 34.10 & 20.52 &  4.00 & 48.50 \\

\rowcolor{highlightblue}
Hybrid
    &  {\underline{69.38}} & {\underline{64.68}} &{\underline{52.50}} & {\underline{70.56}}
    & {\underline{67.42}} & {\underline{63.37} }& \textbf{54.00} & {\underline{73.00}} \\

\rowcolor{highlightblue}
Adaptive
    & \textbf{69.80}& \textbf{65.19} & \textbf{52.61} & \textbf{70.70}
    & \textbf{68.85}& \textbf{64.04} & \textbf{54.00} & \textbf{74.00} \\

\addlinespace[3pt]
\midrule[\heavyrulewidth]
\addlinespace[3pt]

\rowcolor{panelgray}
\multicolumn{9}{c}{\textbf{Qwen-2.5-VL-7B}} \\
\addlinespace[2pt]

Vanilla Base
    & 29.28 & 25.55 & 15.90 & 36.23
    & 44.15 & 39.50 & 12.00 & 48.00 \\

FIFO
    & 26.71 & 23.49 & 17.19 & 30.13
    & 34.79 & 32.07 & 26.00 & 47.50 \\

NaiveRAG
    & 57.48 & 52.02 & 38.26 & 66.59
    & 41.06 & 39.17 & 36.00 & 48.00 \\

Mem0
    & 30.24 & 26.52 & 18.30 & 35.00
    & 43.50 & 41.90 & 38.00 & 50.00 \\

Full-Context
    & 35.72 & 32.21 & 24.58 & 40.32
    & 38.73 & 35.08 & 26.00 & 53.50 \\

MemGPT
    & 52.69 & 47.88 & 32.81 & 66.10
    & 42.31 & 37.20 & 20.00 & 56.50 \\

\cmidrule(r){1-5} \cmidrule(l){6-9}

Full-MM-Context
    & 29.05 & 25.59 & 19.04 & 34.06
    & 39.68 & 36.18 & 26.00 & 52.00 \\

MuRAG
    & 63.97 & 58.95 & 43.90 & 71.05
    & 45.43 & 38.93 & 14.00 & 59.00 \\

UniversalRAG
    & 62.42 & 56.60 & 42.14 & 70.89
    & 44.09 & 37.93 & 14.00 & 56.50 \\

MemVerse
    & 30.88 & 27.36
    & 18.94 & 37.27
    & 34.31 & 27.25 & 6.00 & 57.50 \\

\cmidrule(r){1-5} \cmidrule(l){6-9}

\rowcolor{highlightblue}
Episodic
    & 61.89 & 56.54 & 42.33 & 70.61
    & 48.12 & 43.29
    & 30.00 & 63.50 \\

\rowcolor{highlightblue}
Latent
    & 37.74 & 33.63 & 25.51 & 40.02
    & 33.40 & 19.80 &  4.00 & 50.00 \\

\rowcolor{highlightblue}
Hybrid
    & {\underline{71.27}} & \underline{66.76} & \textbf{54.81} & {\underline{73.29}}
    & \textbf{67.49} & \textbf{63.70} & \textbf{56.00} & {\underline{71.50}} \\

\rowcolor{highlightblue}
Adaptive
    & \textbf{71.55} & {\textbf{67.19}} & {\underline{54.76}} & \textbf{75.28}
    & {\underline{65.56}} & {\underline{61.10}} & {\underline{52.00}} & \textbf{77.50} \\

\bottomrule
\end{tabular}
\end{table*}

\paragraph{Implementation Details.}
We conduct all experiments using Qwen2.5-VL-3B-Instruct and Qwen2.5-VL-7B-Instruct \citep{bai2025qwen25vl} as the inference backbones. For methods requiring LLM-based memory construction, we adopt DeepSeek-V4-Flash~\citep{deepseekai2026deepseekv4}. 
We implement latent memory using LoRA adapters with  $r=256$, applied to the final 4 decoder layers of each backbone. We reserve a small subset of the original test set for outer-loop training and evaluate all methods on the same remaining test samples, and Section~\ref{sec_sensitivity} confirm that this protocol rules out train–test contamination. $\tau_d$ and $\tau_c$ are adaptively calibrated on the training set after the latent memory training phase.
Further implementation details are provided in Appendix.

\subsection{Main Results}
\label{sec:rq_main}

\Cref{tab:main_results} shows Hybrid and Adaptive achieve top scores on every metric across both benchmarks and model sizes. Compared to the strongest baseline, Hybrid improves Judge by 2.98\%/2.24\% on Mem-Gallery, and by 23.0\%/12.5\% on LoCoMo (for 3B/7B backbones). Beyond semantic correctness, the large gains in F1, BLEU-1, and EM indicate answering efficiency and stylistic alignment with memory coordination. 
Comparing individual memory modes reveals a strong complementarity between explicit evidence and implicit experience.
For tasks emphasizing fine-grained details, sufficient evidence retrieval and effective organization within limited context windows are crucial: unimodal retrieval clearly underperforms multimodal retrieval without visual evidence. Our multi-index scoring episodic memory shows a strong foundation for evidence-grounded answering, performing competitively with MuRAG.
While latent memory is less suited for recovering fine-grained details, it surpasses Full-(MM-)Context, proving that native internalization provides a more reliable representation than per-query in-context processing. MemVerse's LLM-based router assigns approximately 65\% of queries to its latent memory branch, it still underperforms our standalone Latent. This degradation reveals the unreliability of internalizing hand-crafted memory storage with loss of global semantic information and highlights the gap between explicit external injection and native model inference utilization.
Adaptive routing further improves performance (e.g., +1.2\%/+2\% Judge on Mem-Gallery/LoCoMo with 7B) by reliably selecting among the three memory modes.



\subsection{Analysis for Task-Specific Memory Preferences}
\label{sec:rq_buckets}

\begin{figure*}[t]
\centering
\noindent

\begin{minipage}[t]{0.635\linewidth}
\vspace{0pt}
\centering

\captionof{table}{
\textbf{Performance across task categories.}
Comparison of Latent, Episodic, and Hybrid across nine
task categories in Mem-Gallery with 7B, highlighting task-dependent
strengths.
}
\label{tab:bucket_results}

\vspace{3pt}

\scriptsize
\setlength{\tabcolsep}{2.5pt}
\renewcommand{\arraystretch}{1.05}

\resizebox{\linewidth}{!}{%
\begin{tabular}{lccccccccc}
\toprule
\textbf{Method}
& \textbf{FR}
& \textbf{MR}
& \textbf{TR}
& \textbf{VR}
& \textbf{TTL}
& \textbf{VS}
& \textbf{KR}
& \textbf{CD}
& \textbf{AR} \\
\midrule

Latent
& 21.10
& 33.97
& 37.97
& 26.80
& 62.38
& 2.08
& 38.94
& 59.62
& \textbf{98.28} \\

Episodic
& \textbf{67.38}
& \textbf{73.09}
& 68.04
& 46.85
& 79.45
& \textbf{73.18}
& 66.83
& 38.46
& 92.24 \\

Hybrid
& 64.54
& 72.52
& \textbf{69.94}
& \textbf{52.03}
& \textbf{84.25}
& 72.53
& \textbf{69.71}
& \textbf{69.23}
& 93.10 \\

\bottomrule
\end{tabular}%
}

\end{minipage}%
\hfill
\begin{minipage}[t]{0.345\linewidth}
\vspace{0pt}
\centering

\includegraphics[
    width=\linewidth
]{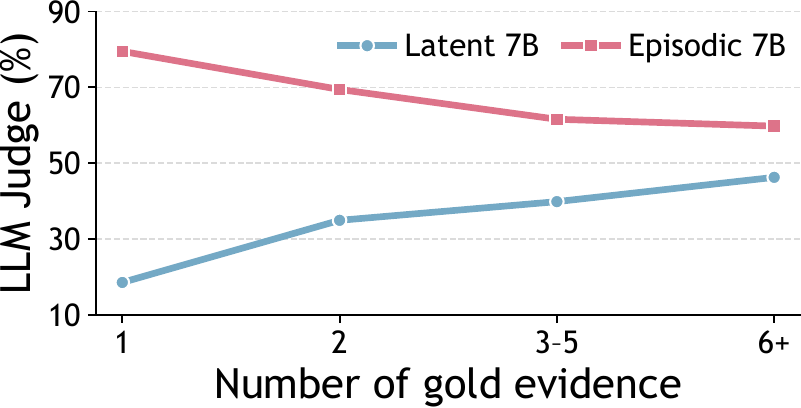}

\vspace{2pt}

\captionof{figure}{
Performance under different gold evidence amounts.
}
\label{fig:counts}

\end{minipage}

\end{figure*}
Following the task taxonomy established by Mem-Gallery~\citep{memgallery2026}, we analyze the performance trade-offs across Latent, Episodic, and Hybrid, as shown in \Cref{tab:bucket_results}, revealing distinct strengths and task-dependent benefits.
For reasoning-intensive tasks including Timing Reasoning (TR, +1.90\%) and Visual Reasoning (VR, +5.18\%), with Test-Time Learning (TTL, +4.81\%) which demands rapid knowledge internalization, Hybrid and Latent show marked gains. These tasks benefit from compacting long-term visual facts to facilitate integrated reasoning.
For Knowledge Resolution (KR, +2.88\%) and Contradiction Detection (CD, +30.77\%), which call for the comprehensive awareness, are substantially augmented by latent implication.

Conversely, for fine-grained, evidence-dependent tasks including Fact Retrieval (FR), Multi-Entity Reasoning (MR), and Visual Search (VS), pure Episodic maintains an advantage  (+0.71\%/+3.52\%/+4.81\%). These tasks rely heavily on grounded multimodal evidence without abstract loss, where Latent fall short (only 2.08\% on VS and 21.10\% on FR). 
For Answer Refusal (AR), which evaluates the ability to refrain from answering, Latent memory aids in recognizing broader context gaps, enabling more conservative refusal behaviors.

In addition, we evaluate performance relative to the number of required evidence in \Cref{fig:counts}. As gold evidence scales, Episodic degrades due to noise accumulation and retrieval precision loss. In contrast, Latent rises as inherently well-suited for global modeling.
Overall, these task-dependent memory preferences highlight the need for memory integration and dynamic routing.

\begin{table*}[h]
\centering

\begin{minipage}[t]{0.56\textwidth}
\vspace{0pt}
\centering

\includegraphics[width=\linewidth]{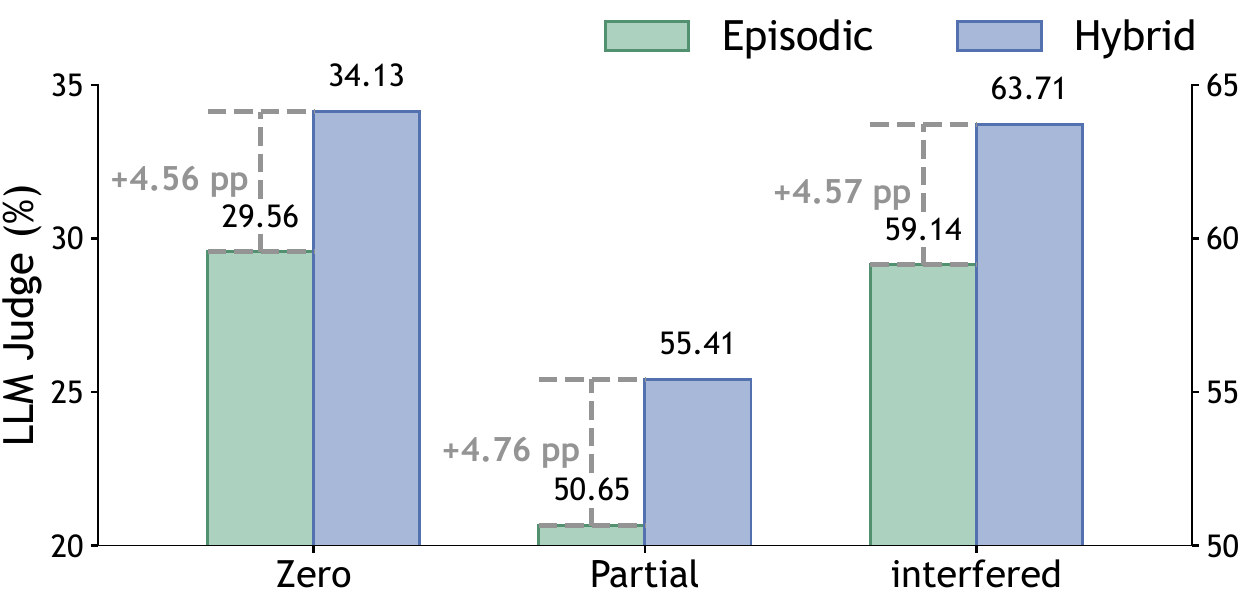}

\captionof{figure}{Comparison of Episodic and Hybrid under insufficient(zero and partial) and interfered retrieval.
}
\label{fig:retrieval_gap}
\end{minipage}
\hfill
\begin{minipage}[t]{0.41\textwidth}
\vspace{0pt}
\centering
\small

\captionof{table}{
Training-label leakage analysis. \textit{Seen(\%)} is the fraction of test gold answers occurring in same-scene training labels. \textit{Unseen@EM(\%)} is the fraction of exact-match predictions whose gold answer is absent from the training labels.}

\label{tab:semantic_overlap}

\setlength{\tabcolsep}{3pt}
\resizebox{\linewidth}{!}{%
\begin{tabular}{l S[table-format=3.2] S[table-format=3.2]}
\toprule[1.0pt]
Answer Interface & {Seen} & {Unseen@EM} \\
\midrule[0.5pt]
Open Semantic Answer & 0.00 & 100.00 \\
Local Evidence Pointer & 15.64 & 95.04 \\
Shared Canonical Label & 52.86 & 37.16 \\
Normalized Scalar & 34.83 & 55.81 \\
\bottomrule[1.0pt]
\end{tabular}
}








\end{minipage}

\end{table*}








\subsection{Latent Memory Compensates for Retrieval Failures}
\label{sec:rq_outer}
We examine latent memory as a complementary mechanism under primary Episodic failures caused by misaligned or incomplete retrievals.
We compare Episodic with Hybrid under constructed imperfect-retrieval conditions: insufficient retrieval(zero and partial clue recall) and retrieval interference(wrong item ranks top-1), as \Cref{fig:retrieval_gap}.
Under partial and zero recall, Hybrid  achieves relative gains of 9.5\% and 15.2\%, suggesting effective information completion by Latent.
For retrieval interference, Hybrid improves Judge from 59.1\% to 63.7\%, highlighting its error correction capability.
Together, these results demonstrate that latent memory effectively bridges the retrieval blind spots of hand-engineered strategies in long-tail scenarios.

\subsection{Sensitivity to Hyperparameters}
\label{sec_sensitivity}
\begin{figure*}[t]
\centering
\noindent

\begin{minipage}[t]{0.44\linewidth}
\vspace{0pt}
\centering

\captionof{table}{
Average computational cost including Ingest latency per scene, QA latency and token usage per query.
}
\label{tab:locomo_costs}

\vspace{3pt}

\footnotesize
\setlength{\tabcolsep}{3pt}
\renewcommand{\arraystretch}{1.05}

\resizebox{\linewidth}{!}{%
\begin{tabular}{
  l
  S[table-format=4.4]
  S[table-format=2.2]
  S[table-format=4.2]
}
\toprule
& \multicolumn{2}{c}{\textbf{Latency (s)}}
& {\textbf{Token Usage}} \\
\cmidrule(lr){2-3}
\cmidrule(lr){4-4}

\textbf{Method}
& {\textbf{Ingest / scene}}
& {\textbf{QA / q}}
& {\textbf{Input tok / q}} \\

\midrule
Mem0            & 1504.98 & 0.64  & 679.13  \\
Full-Context    & 0.00    & 1.21  & 4569.82 \\

\midrule
Full-MM-Context & 0.00 & 21.00 & 28416.82 \\
MuRAG           & 136.62  & 2.64  & 3274.82 \\
UniversalRAG    & 138.77 & 3.82 & 2574.12 \\
MemVerse        & 2663.50 & 19.94 & 2866.33 \\

\midrule
\rowcolor{highlightgray}
Latent
& 3244.72
& 0.68
& 510.84 \\
\rowcolor{highlightgray}
Episodic
& 1958.61
& 10.91
& 2626.35 \\

\bottomrule
\end{tabular}%
}

\end{minipage}%
\hfill
\begin{minipage}[t]{0.54\linewidth}
\vspace{0pt}
\centering

\includegraphics[
    width=\linewidth
]{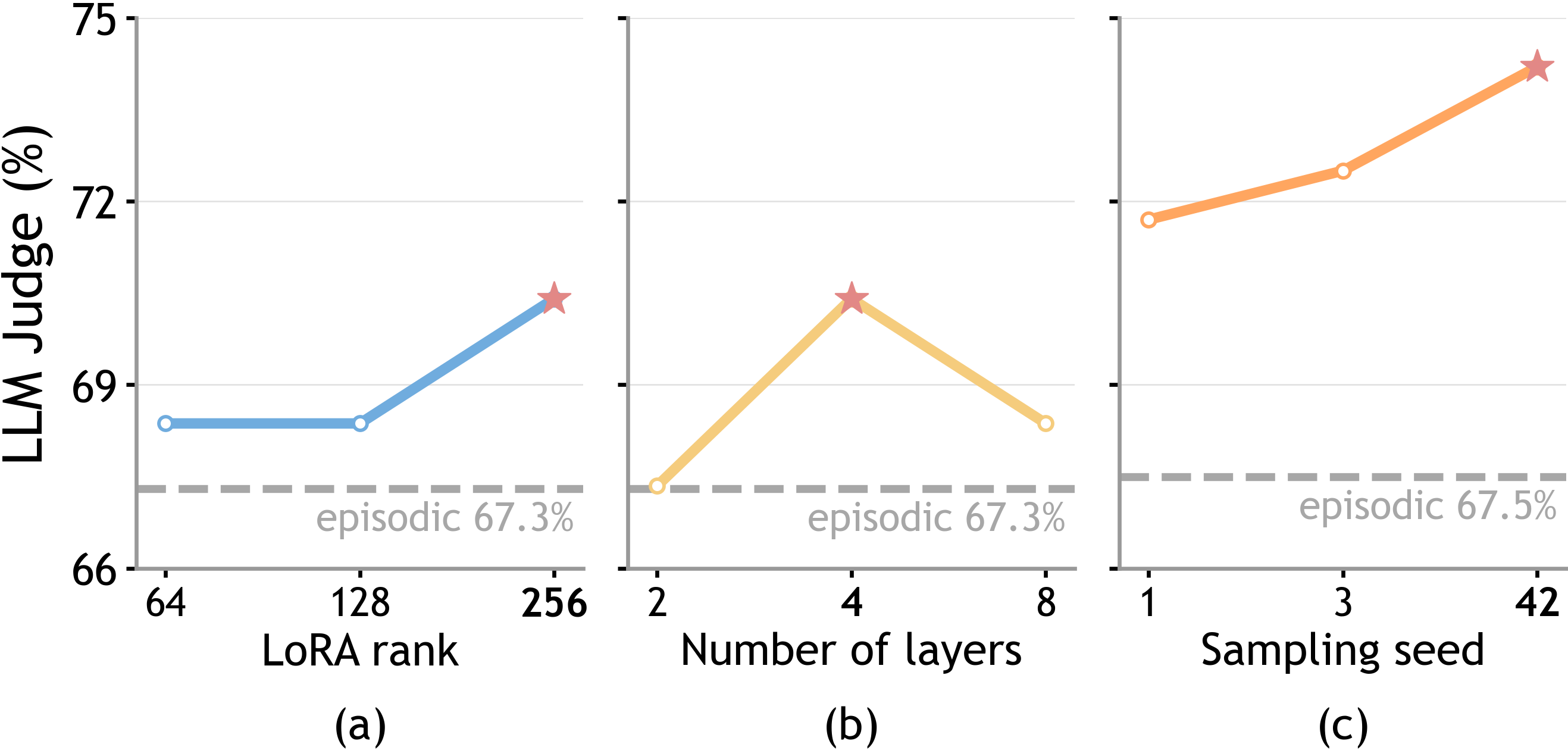}

\vspace{2pt}

\captionof{figure}{
Hyperparameter sensitivity of Hybrid Memory on a Mem-Gallery scenario across varying
(a) LoRA rank,
(b) number of adapted layers,
and (c) training-set sampling seed.
}
\label{fig:hybrid_sweeps}

\end{minipage}

\end{figure*}

First, sweeping LoRA rank (\Cref{fig:hybrid_sweeps}a) shows that while low-rank adapters yield initial gains, higher capacity better supports knowledge acquisition and joint use.
Meanwhile, evaluating injection depth across the final $n$ layers (\Cref{fig:hybrid_sweeps}b) reveals peak accuracy at the top $n=4$ layers, indicating adaptation is best concentrated in layers that efficiently exploit latent memory while preserving core backbone features.
Second, to rule out data selection sensitivity, we evaluate three resampled training sets on a fixed test set (\Cref{fig:hybrid_sweeps}c), achieving consistent performance.
Finally, to further exclude sole pattern memorization despite our conservative learning rate and limited optimization steps, we analyze train--test answer overlap across open-answer interfaces. According to Table~\ref{tab:semantic_overlap}, the two more open-ended and evidence-specific interfaces provide little opportunity for direct answer reuse. For the two more confounded interfaces, Shared Canonical Labels and Normalized Scalars, although they have a higher answer overlap, the most frequent same-scene training answer achieves only 8.45\% and 11.24\% EM, far below the Hybrid's 71.12\% and 48.31\%. Overall, these results confirm that latent memory indeed captures global implicit representations aligned with target outputs, rather than merely fitting training-set patterns.

\subsection{Cost Analysis}
\label{subsec:cost_analysis}

Table~\ref{tab:locomo_costs} compares construction and query costs across methods on Mem-Gallery with Qwen2.5-VL-7B. 
While structured memory incurs upfront ingest overhead, it significantly optimizes test-time inference. 
Specifically, Episodic memory reduces input tokens by $42.5\%$, with pure inference taking $\sim0.9$s. 
Latent requires higher ingest latency due to meta-learning, but achieves extreme test-time efficiency ($0.68$s, $510.84$ tokens) with superior performance over Full-Context. 
Overall, our approaches trade initial ingest overhead for optimal query efficiency and superior performance.

\section{Conclusion}
\label{sec:conclusion}

In this work, we characterize the complementary trade-offs of episodic and latent memory in multimodal agents and present a dual-level test-time adaptation framework. The framework couples a multi-indexed episodic memory layer for evidence grounding with a native latent adapter optimized via test-time adaptation for semantic internalization. To mitigate cross-branch interference, we further introduce a training-free adaptive routing mechanism driven by two proxies. Extensive experiments on multimodal long-term benchmarks demonstrate that the hybrid framework significantly outperforms single-paradigm baselines with superior robustness and efficiency. Moreover, different tasks show marked variations in their memory mechanism preferences across scenarios.

\paragraph{Limitation and Future works.}
\textbf{(1) Task Scenarios:} Constrained by available standard benchmarks that lean toward episodic memory, the framework can be extended to scenarios with higher procedural memory demands.
\textbf{(2) Routing Strategy:} To minimize overhead, our router relies on low-cost proxies; lightweight trainable parameters can be integrated to potentially improve precision.
\textbf{(3) Memory Evolution:} While both memories are constructed post-trajectory with fixed distributions, reflective collaborative updates can be incorporated to enable dynamic evolution.

\bibliography{references}
\bibliographystyle{iclr2027_conference}
\clearpage
\appendix
\section{Notation Summary}

Table~\ref{tab:notations} summarizes the main symbols employed throughout this paper. The notation follows the dual-level memory structure of our framework. Episodic Memory uses raw records, derived multi-granularities, and query relevance channels for historical retrieval. Latent Memory uses adaptive LoRA states for knowledge internalization. Adaptive Routing Proxy examines token distributions and gating functions to coordinate memory activation.
\begin{table}[htbp]
\centering
\caption{Summary of mathematical notations and definitions} 
\label{tab:notations}
\small
\renewcommand{\arraystretch}{1.2} 
\begin{tabular}{p{0.22\textwidth} p{0.24\textwidth} p{0.45\textwidth}}
\toprule
\textbf{Symbol} & \textbf{Definition} & \textbf{Description} \\
\midrule
\multicolumn{3}{l}{\textbf{\textit{Episodic Memory}}} \\
\addlinespace[2pt]
\(\mathcal{R}, r\) & Record set \& item & Set of raw historical records (\(\mathcal{R}\)) and a single record (\(r \in \mathcal{R}\)). \\
\(\mathcal{M}\) & Derived memories & The multi-granularity memory system \(\{\mathcal{M}_{sum}, \mathcal{M}_{evt}, \mathcal{M}_{sem}, \mathcal{M}_{res}\}\). \\
\(q\) & Input query & Current input query containing text and optional images. \\
\(s_{(\cdot)}, \lambda_{(\cdot)}\) & Scores \& weights & Relevance scores and weight coefficients for retrieval channels. \\
\(S(q, r), \mathcal{R}_q\) & Score \& evidence & Combined score between \(q\) and \(r\), and top-\(K\) selected evidence set. \\
\addlinespace[4pt]
\multicolumn{3}{l}{\textbf{\textit{Latent Memory}}} \\
\addlinespace[2pt]
\(\theta, \phi\) & Backbone \& adapter & Parameters of frozen base model (\(\theta\)) and implicit LoRA adapter (\(\phi\)). \\
\(\mathcal{L}_{inner}, \mathcal{L}_{outer}\) & Loss functions & Inner-loop NTP loss for internalization and outer-loop answer loss. \\
\(C_r, \mathcal{D}_{align}\) & Chunks \& support set & Historical record chunks (\(C_r\)) and small QA support set (\(\mathcal{D}_{align}\)). \\
\(\phi_t, \phi_t^{\prime}, \eta\) & States \& learning rate & Adapter initial state, adapted state at iteration \(t\), and outer-loop learning rate. \\
\addlinespace[4pt]
\multicolumn{3}{l}{\textbf{\textit{Adaptive Routing Proxy}}} \\
\addlinespace[2pt]
\(d(q, \phi)\) & JSD proxy metric & First-stage proxy measuring query sensitivity to implicit memory. \\
\(P_{\theta}^{(1)}, P_{\theta,\phi}^{(1)}\) & Token distributions & First-generation token distributions of base and combined models. \\
\(c_L(q, \phi), T, \hat{a}_t\) & Confidence \& sequence & Second-stage confidence, sequence length \(T\), and generated token \(\hat{a}_t\). \\
\(\tau_d, \tau_c\) & Thresholds & JSD threshold (\(\tau_d\)) and confidence threshold (\(\tau_c\)). \\
\(g_L, g_{\mathcal{E}}\) & Gating functions & Switching gating functions for implicit memory and explicit retrieval. \\
\bottomrule
\end{tabular}

\end{table}
\section{Experimental Protocol}
\subsection{Benchmarks}
\label{sec:appendix-benchmarks}

To evaluate long-term memory in multimodal conversational agents, we use two complementary benchmarks, Mem-Gallery and the LoCoMo-0--9 subset of MemoryBench.

\paragraph{\textbf{Mem-Gallery} \citep{memgallery2026}} evaluates multimodal long-term conversational memory in multimodal large language model (MLLM) agents. It contains long-horizon, multi-session conversations grounded in both visual and textual information. The benchmark therefore tests whether an agent can retain and organize cross-modal evidence over time. Its evaluation protocol considers three aspects of memory. The first is \textbf{memory extraction and test-time adaptation}, which measures whether newly observed information can be retained and used in later interactions. The second is \textbf{memory reasoning}, which evaluates reasoning over retrieved textual and visual memories. The third is \textbf{memory knowledge management}, which covers knowledge resolution, conflict detection, and answer refusal when the stored evidence is insufficient.

\paragraph{\textbf{MemoryBench (LoCoMo-0--9)} \citep{memorybench2025}} evaluates memory and continual learning in LLM systems. It provides a user-feedback simulation framework that covers multiple domains, languages, and task formats. We use its LoCoMo-0--9 subset, which consists of very long, multi-session conversations with temporal and multimodal dependencies. Each conversation contains approximately 600 turns and 16K tokens on average and spans up to 32 sessions. These tasks require models to retrieve and reason over information accumulated throughout extended interaction histories.

\subsection{Data Splits.}
Mem-Gallery contains 1,711 QA pairs across 20 scenarios and does not provide an official training split. We therefore construct a fixed split independently within each scenario using random seed 42, targeting approximately 40\% of the QA pairs for training and 60\% for testing. When necessary, the split is slightly adjusted to ensure that every task category within a scenario is represented in the test set. This results in 629 training and 1,082 test questions. The same split is used across all methods and model scales, and all reported Mem-Gallery results are evaluated on the 1,082-question test set. For LoCoMo, we follow the official experimental setting with 200 training and 50 test questions, with evaluation performed on the official test set. Detailed statistics of the Mem-Gallery split are reported in Table~\ref{tab:memgallery_split}.

\begin{table*}[t]
\centering
\small
\caption{Train--test split statistics for the 20 scenarios in Mem-Gallery. The split is constructed independently within each scenario with random seed 42.}
\label{tab:memgallery_split}
\begin{tabular}{
>{\raggedright\arraybackslash}p{0.66\textwidth}
>{\centering\arraybackslash}p{0.07\textwidth}
>{\centering\arraybackslash}p{0.07\textwidth}
>{\centering\arraybackslash}p{0.07\textwidth}}
\toprule
\textbf{Scenario} & \textbf{Train} & \textbf{Test} & \textbf{Total} \\
\midrule
AI, Robotics, Automation, Future Tech & 21 & 36 & 57 \\
Academic, Animal, Pet Research, Life & 28 & 49 & 77 \\
Architecture, Art, Culture, Exhibition, Technology & 26 & 48 & 74 \\
Astronomy, Physics, Scientific Experiments, Cosmology & 21 & 33 & 54 \\
Baking, Dessert, Daily Life, Skill & 41 & 70 & 111 \\
Dog Behavior, Research, Academic, Life & 24 & 39 & 63 \\
Education, Career, Research, Lifestyle & 38 & 68 & 106 \\
Entrepreneurship, Blockchain, Economics, Logistics, Nature & 33 & 57 & 90 \\
Fashion, Personal Care, Lifestyle, Shopping & 22 & 39 & 61 \\
Global Travel, Cultural Sightseeing & 31 & 56 & 87 \\
Global Travel, Sustainable Fashion, Design & 29 & 49 & 78 \\
Home, Health, Lifestyle, Product & 25 & 49 & 74 \\
Home Repair, Maintenance, Cleaning & 26 & 42 & 68 \\
Landscape, Travel, Architecture, Nature & 34 & 58 & 92 \\
Music, Dance, Theater, Performance, Learning & 19 & 38 & 57 \\
Nature, Economics, Programming, Student Life & 52 & 84 & 136 \\
Parenting, Commuting, Hobbies, Travel Gear & 30 & 52 & 82 \\
Python, Botany, AI, Student Life & 78 & 127 & 205 \\
Real Estate, Home Decor, DIY, Lifestyle & 27 & 48 & 75 \\
Technology, Ethics, Future Society & 24 & 40 & 64 \\
\midrule
\textbf{Total} & \textbf{629} & \textbf{1,082} & \textbf{1,711} \\
\bottomrule
\end{tabular}
\end{table*}
\subsection{Evaluation Metrics}
\label{app:metrics}
We evaluate the quality of generated answers using three reference-based lexical metrics and an LLM-based evaluation metric.

\paragraph{\textbf{Token-level F1.}}
Token-level F1 measures the overlap between the generated answer and the reference answer at the token level \citep{rajpurkar-etal-2016-squad}. Let $P$ and $R$ denote token-level precision and recall, respectively:
\begin{equation}
P = \frac{N_{\mathrm{overlap}}}{N_{\mathrm{pred}}},
\qquad
R = \frac{N_{\mathrm{overlap}}}{N_{\mathrm{ref}}},
\end{equation}
where $N_{\mathrm{overlap}}$ is the number of overlapping tokens between the prediction and reference, and $N_{\mathrm{pred}}$ and $N_{\mathrm{ref}}$ are the numbers of tokens in the prediction and reference, respectively. The F1 score is then computed as
\begin{equation}
\mathrm{F1} = \frac{2PR}{P+R}.
\end{equation}
This metric assigns partial credit when the generated answer contains only part of the reference information.

\paragraph{\textbf{BLEU-1.}}
BLEU-1 measures unigram-level precision between the generated and reference answers \citep{papineni-etal-2002-bleu}. It evaluates how many tokens in the generated answer also appear in the reference, together with the standard brevity penalty that discourages overly short predictions. Compared with Token-level F1, BLEU-1 emphasizes precision of lexical matches rather than balancing precision and recall.

\paragraph{\textbf{Exact Match.}}
Exact Match (EM) measures whether the generated answer exactly matches the reference answer after normalization \citep{rajpurkar-etal-2016-squad}. Each prediction receives a score of 1 if it matches the reference and 0 otherwise, and the final EM score is averaged over all evaluation samples. EM therefore provides a strict measure of answer correctness but does not give partial credit to semantically correct answers with different surface forms.

\paragraph{\textbf{LLM-as-a-Judge.}}
Since lexical matching metrics may underestimate answers that are semantically correct but expressed differently from the reference, we additionally employ an LLM-based evaluator. We follow the official evaluation prompt provided by Mem-Gallery \citep{memgallery2026} and use DeepSeek-V4-Flash as the judge model. Given the question, reference answer, and model-generated answer, the judge evaluates the response according to its correctness, completeness, consistency with the reference, and the presence of incorrect or hallucinated information.

For each sample, the judge assigns one of five discrete scores,
\begin{equation}
s \in {0, 0.25, 0.5, 0.75, 1},
\end{equation}
where $0$ denotes an incorrect answer, $0.25$ a largely incorrect or tangential answer, $0.5$ a partially correct or incomplete answer, $0.75$ a largely correct answer with minor omissions or imperfections, and $1$ a fully correct and precise answer. The final LLM-as-a-Judge score is computed by averaging the scores over all evaluation samples.

\section{Implementation Details}
\label{app:implementation}

\subsection{Model and Inference Settings.}
We use Qwen2.5-VL-3B-Instruct and Qwen2.5-VL-7B-Instruct \citep{bai2025qwen25vl} as the inference backbones. For methods requiring LLM-based memory construction, we use DeepSeek-V4-Flash~\citep{deepseekai2026deepseekv4}. All LLM-based generation and memory-construction calls use a temperature of 0. For each benchmark, all methods use the same official question-answering prompt and generation settings; the primary method-specific difference lies in the memory provided to the inference backbone. The common inference settings are summarized in Table~\ref{tab:inference_settings}.

\begin{table}[t]
\centering
\small
\caption{Model and inference settings used in our experiments.}
\label{tab:inference_settings}
\begin{tabular}{
>{\raggedright\arraybackslash}p{0.25\linewidth}
>{\raggedright\arraybackslash}p{0.31\linewidth}
>{\raggedright\arraybackslash}p{0.31\linewidth}}
\toprule
Setting & Mem-Gallery & LoCoMo \\
\midrule
Inference backbone & Qwen2.5-VL-3B/7B-Instruct & Qwen2.5-VL-3B/7B-Instruct \\
Memory construction & DeepSeek-V4-Flash (when required) & DeepSeek-V4-Flash (when required) \\
Temperature & 0 & 0 \\
Retrieval top-$K$ & 10 & 10 \\
Max. output tokens & 256 & 8,192 \\
QA prompt & Official & Official \\
\bottomrule
\end{tabular}
\end{table}

\subsection{Baseline Implementations.}
We follow the implementations provided by Mem-Gallery whenever applicable and use the unified inference settings described above. Below, we briefly describe the memory construction and retrieval mechanisms of the evaluated baselines.

\paragraph{Vanilla Base.}
Vanilla Base does not maintain any external memory. The backbone model directly answers each query without retrieving or incorporating information from previous interactions, serving as the no-memory reference.

\paragraph{Full Memory (Text).}
Full Memory (Text) stores the complete textual conversation history without additional memory compression or relevance-based retrieval. At inference time, the stored memories are concatenated and provided directly to the backbone model, with truncation applied when the accumulated context exceeds the available context budget.

\paragraph{First-in-first-out (FIFO).}
FIFO maintains a short-term memory by retaining the most recent conversational content. When the memory capacity is exceeded, older entries are discarded following a first-in-first-out strategy, and the remaining recent memories are concatenated as context for answering the current query.

\paragraph{NaiveRAG.}
NaiveRAG stores conversation turns as textual memory entries and retrieves relevant memories according to the semantic similarity between the query and text embeddings. The top-$K$ retrieved memories are then provided to the backbone model as additional context.

\paragraph{MemGPT.}
MemGPT adopts a hierarchical memory mechanism consisting of a working context, a FIFO-based queue, recursive summarization, recall storage, and archival storage. Its memory management mechanism determines when conversational information is retained in the working context, summarized, archived, or retrieved for subsequent inference.

\paragraph{Full Memory (Multimodal).}
Full Memory (Multimodal) stores the complete multimodal conversation history, including both textual content and raw images. At inference time, the accumulated multimodal memory is directly provided to the backbone model without relevance-based retrieval. When the memory exceeds the available text-token or image budget, older entries are truncated while more recent memories are retained.

\paragraph{MuRAG.}
MuRAG stores both textual and visual conversational memories and represents them using multimodal embeddings. Given a query, relevant text--image memories are retrieved according to multimodal embedding similarity, and the top-$K$ retrieved memories are provided to the backbone model for response generation.

\paragraph{UniversalRAG.}
UniversalRAG stores and encodes both textual and visual memories. Given a query, an LLM-based router first determines the appropriate retrieval modality, including no retrieval, textual retrieval, or visual retrieval, after which the corresponding retrieval procedure is performed and the retrieved memories are provided to the backbone model.

\paragraph{Mem0.}
Mem0 constructs textual memories by first using an LLM to extract relatively stable user-specific facts from each conversation segment. Each extracted fact is then compared against existing memories to identify potential redundancy or conflicts, based on which the memory is added, updated, deleted, or retained. At inference time, the memory store retrieves the top-$K$ entries most relevant to the current query and provides them as context for downstream question answering. 

\paragraph{MemVerse.}
Following the original MemVerse design, we enable both retrieval-based and parametric memory pathways. Conversational content is first processed by the MemVerse service to construct external memory, while a distilled Qwen2.5-1.5B-Instruct model is used to construct the parametric memory as a complementary source of reference information. At inference time, an LLM-based routing mechanism selects which memory pathway to use for answering the current query.

\subsection{Training Configuration.}
We train the latent memory through LoRA adaptation of Qwen2.5-VL-3B-Instruct and Qwen2.5-VL-7B-Instruct. LoRA modules are applied only to the attention projections in the last four Transformer layers, while all non-LoRA parameters remain frozen. The complete training configuration is summarized in Table~\ref{tab:training_settings}.

\begin{table}[t]
\centering
\small
\caption{Training configuration for latent memory optimization.}
\label{tab:training_settings}
\begin{tabular}{
>{\raggedright\arraybackslash}p{0.43\linewidth}
>{\raggedright\arraybackslash}p{0.46\linewidth}}
\toprule
Hyperparameter & Value \\
\midrule
Base model & Qwen2.5-VL-3B/7B-Instruct \\
LoRA target layers & Last 4 Transformer layers \\
LoRA target modules & \texttt{q\_proj}, \texttt{k\_proj}, \texttt{v\_proj}, \texttt{o\_proj} \\
LoRA rank $r$ & 256 \\
LoRA $\alpha$ & 256 \\
LoRA dropout & 0 \\
LoRA $A$ initialization & Kaiming uniform \\
LoRA $B$ initialization & Zero \\
Inner steps & 4 \\
Outer steps & 5 \\
Inner learning rate & $1\times10^{-4}$ \\
Outer learning rate & $5\times10^{-5}$ \\
Weight decay & 0.01 \\
Maximum gradient norm & 1.0 \\
Context chunk length & 256 tokens \\
Chunks per micro-batch & 4 \\
Random seed & 42 \\
Hardware & 7$\times$ NVIDIA RTX A6000 \\
\bottomrule
\end{tabular}
\end{table}

\subsection{Training Procedure.}
Training follows a first-order bilevel optimization procedure. At each outer step, the LoRA parameters are first adapted to the dialogue context through inner-loop updates. The adapted model is then evaluated on fixed, precomputed multimodal QA samples. We compute an answer-only cross-entropy loss, where the system and user prompt tokens are masked and only the answer tokens contribute to the training objective. For the outer update, the QA gradient is computed at the inner-adapted parameters. We then restore the LoRA parameters to their values before inner adaptation and apply this gradient to update the outer LoRA initialization. This first-order formulation does not backpropagate through the inner optimization trajectory. A new inner optimizer is instantiated for each outer step, whereas the outer optimizer is maintained across outer steps.

Dialogue contexts used for inner adaptation are divided into 256-token chunks. The multimodal QA samples used for outer-loop supervision are precomputed using our retrieval method on the corresponding training split and remain fixed throughout training. The official benchmark prompts are used for QA supervision.

\section{Additional Experimental Results}

\subsection{Sensitivity to Model Seed}
\label{sec:model_seed}

\begin{table}[t]
\centering
\small
\caption{Sensitivity of Hybrid Memory to different model seeds in Scenario 1(Academic). The gain is measured against Episodic Memory (0.673).}
\label{tab:model_seed}
\begin{tabular}{ccc}
\toprule
Model Seed & Performance & Gain over Episodic \\
\midrule
41 & 0.730 & +0.057 \\
42 & 0.704 & +0.031 \\
43 & 0.689 & +0.016 \\
\bottomrule
\end{tabular}
\end{table}

Table~\ref{tab:model_seed} evaluates the sensitivity of Hybrid Memory to the random seed used to initialize the LoRA matrix \(A\), while keeping the training and test sets fixed, with seed 42 used in our main experiments. The performance varies only moderately across different initializations, ranging from 0.689 to 0.730. Although seed 41 achieves the highest score, all three settings consistently outperform Pure Episodic Memory, with gains of 0.016--0.057. These results show that the improvement brought by Hybrid Memory is consistently preserved across different LoRA initializations.

\subsection{Effect of Task-Specific Prompting}
\label{sec:prompt_ablation_full}

Mem-Gallery provides task-specific prompts for several question categories, specifying the expected answering behavior and output format. Representative examples for AR, CD, and VS are shown below.

\begin{quote}
\small
\textbf{AR:} ``Provide your answer based on the information in the conversation. Only if the information about the question is not present in the conversation, reply with: `Not mentioned.' ''

\textbf{CD:} ``Please check whether this information conflicts with the conversation, and reply strictly with either `Yes.' or `No.' ''

\textbf{VS:} ``Return the image\_id of the image(s). If there are multiple images, sort them in ascending order and separate them by commas. Format example: `D2:IMG\_003, D2:IMG\_010, D10:IMG\_002' (for format reference only).''
\end{quote}

These prompts therefore provide explicit task-level guidance beyond the retrieved evidence itself. To examine how such guidance affects memory utilization and task-specific answering behavior, we remove the task-specific instructions and evaluate both Episodic Memory and Hybrid Memory on Scenario 1 (Academic). 

\begin{table}[t]
\centering
\small
\caption{Full results of the task-specific prompting ablation on Scenario 1 (Academic). \(\Delta\) denotes the LLM-Judge improvement of Hybrid Memory over Episodic Memory under the same prompting setting.}
\label{tab:prompt_ablation_full}
\setlength{\tabcolsep}{6pt}
\begin{tabular}{ll S[table-format=1.3] S[table-format=1.3] S[table-format=1.3] S[table-format=1.3] l}
\toprule[1.0pt]
Memory & Prompt & {F1} & {BLEU-1} & {EM} & {LLM-Judge} & {\(\Delta\)} \\
\midrule[0.5pt]
Episodic & w/ task prompt
& 0.585 & 0.538 & 0.408 & {\underline{0.673}} & -- \\
Hybrid & w/ task prompt
& {\textbf{0.703}} & {\textbf{0.668}} & {\textbf{0.592}} & {\textbf{0.704}} & +0.031 \\
\midrule[0.5pt]
Episodic & w/o task prompt
& 0.435 & 0.380 & 0.265 & 0.612 & -- \\
Hybrid & w/o task prompt
& {\underline{0.615}} & {\underline{0.580}} & {\underline{0.510}} & 0.658 & +0.046 \\
\bottomrule[1.0pt]
\end{tabular}
\end{table}

According to table~\ref{tab:prompt_ablation_full}, Task-specific prompting substantially improves the Episodic baseline. Compared with the setting without task-specific instructions, its F1 increases from 0.435 to 0.585, EM from 0.265 to 0.408, and LLM-Judge from 0.612 to 0.673. This confirms that the explicit instructions provide strong guidance on how to interpret the task and produce the expected response format. Hybrid Memory also benefits from these prompts, but the improvement is smaller. Its F1 increases from 0.615 to 0.703 and its LLM-Judge score from 0.658 to 0.704. 

More importantly, Hybrid Memory consistently outperforms Episodic Memory under both prompting settings, with a larger advantage when task-specific instructions are removed. Notably, even without such prompts, Hybrid Memory already surpasses the prompted Episodic baseline on F1, BLEU-1, and EM. These results suggest that task-specific prompts help Episodic Memory by explicitly specifying how retrieved evidence should be mapped to task-appropriate answers, whereas parametric memory can implicitly acquire part of this behavior during training. Consequently, Hybrid Memory relies less on explicit task guidance to utilize or organize evidence.

\subsection{Training-label Leakage Analysis}
\label{sec:model_seed}
Table~\ref{tab:appendix_open_category} gives the complete category-level analysis for the seven open-answer categories, aggregated over all 20 scenes. AR and CD are omitted because AR has the structurally fixed answer \textit{Not mentioned} and CD is a closed Yes/No task. The first three columns assess whether the results could be explained by reusing outer-training answer labels. \textit{Seen} is the fraction of test reference answers that occur among same-scene training labels. \textit{Mode EM} is an intentionally favorable label-prior baseline that knows both the scene and category but always returns the most frequent corresponding training answer. \textit{Unseen@EM} measures the proportion of exact Hybrid Memory predictions whose gold answer is absent from the same-scene training labels. The final three columns report the change from Episodic Memory to Hybrid Memory on the complete test set of each category. 

Among the seven open-answer categories, FR and MR provide the clearest evidence against training-label leakage. Their test-label coverage and label-mode baselines are zero or close to zero, yet 89.58\% and 100.00\% of the exact Hybrid Memory hits correspond to gold answers that never appear in the same-scene training labels. TR shows a similar pattern from a different perspective. Although only 27.85\% of its test answers are covered by the training labels, Hybrid Memory still achieves substantial gains across all three semantic metrics. KR and VR further show that train-test label overlap does not by itself account for the observed improvements, since the corresponding label-mode baselines remain much lower than the gains obtained by Hybrid Memory.

TTL provides a complementary case because its category-recognition nature naturally leads to repeated entity labels across splits. Even in this setting, the label-mode baseline reaches only 14.42\% EM, which is far below the improvement achieved by Hybrid Memory. VS has a similarly structured answer space, where answers are image identifiers that can naturally recur across questions. Such label overlap alone does not imply a correct prediction, since selecting the appropriate image ID still requires understanding the query and the corresponding visual evidence. Its negative performance changes further show that reusable answer labels alone are insufficient to produce gains. Together, these results indicate that the improvements of Hybrid Memory cannot be explained by simply selecting or repeating training labels. Instead, it depends on recovering and interpreting information relevant to each test query.

\begin{table*}[t]
\centering
\caption{Complete open-answer analysis by category. Performance changes are percentage-point differences (Hybrid Memory\(-\)Episodic Memory) on the complete category test set. AR and CD are excluded.}
\label{tab:appendix_open_category}
\resizebox{\textwidth}{!}{%
\begin{tabular}{l S[table-format=3.2] S[table-format=3.2] S[table-format=3.2] S[table-format=+2.2] S[table-format=+2.2] S[table-format=+2.2]}
\toprule[1.0pt]
Category & {Seen (\%)} & {Mode EM (\%)} & {Unseen@EM (\%)} & {\(\Delta\)F1} & {\(\Delta\)BLEU-1} & {\(\Delta\)EM} \\
\midrule[0.5pt]
FR & 4.25 & 0.00 & 89.58 & +0.84 & +1.27 & +3.55 \\
KR & 11.54 & 7.69 & 66.67 & +5.94 & +5.18 & +3.85 \\
MR & 0.00 & 0.00 & 100.00 & +2.94 & +2.92 & +3.05 \\
TR & 27.85 & 21.52 & 59.52 & +15.69 & +15.54 & +17.72 \\
TTL & 74.04 & 14.42 & 22.16 & +28.15 & +33.57 & +40.38 \\
VR & 35.14 & 12.61 & 46.00 & +6.19 & +6.34 & +8.11 \\
VS & 16.67 & 4.17 & 94.83 & -0.57 & -1.94 & +0.00 \\
\bottomrule[1.0pt]
\end{tabular}%
}
\end{table*}

\subsection{Effect of Evidence-Conditioned Outer-Loop Supervision.}
\label{app:evidence_conditioned_outer}

To further investigate how evidence affects latent memory, we compare two outer-loop training modes on 133 held-out questions from three scenes of Mem-Gallery: Episodic-Outer, which receives retrieved evidence in its outer-loop QA prompt, and QA-only Outer, which is supervised with the Direct-QA prompt alone. Both variants are evaluated on exactly the same frozen retrieved evidence. Importantly, this evidence is query-retrieved episodic memory rather than the benchmark gold clue, and it can be incomplete or irrelevant.

\begin{table}[t]
\centering
\small
\setlength{\tabcolsep}{4pt}
\begin{tabular}{ll S[table-format=2.2] S[table-format=2.2] S[table-format=2.2] S[table-format=2.2]}
\toprule[1.0pt]
Outer-loop supervision & Method & {F1} & {BLEU-1} & {EM} & {LLM Judge} \\
\midrule[0.5pt]
QA-only Outer & Latent & {\textbf{44.00}} & {\textbf{41.44}} & {\textbf{32.33}} & {\textbf{44.36}} \\
Episodic-Outer & Latent & 40.50 & 37.01 & 29.32 & 43.05 \\
\midrule[0.5pt]
QA-only Outer & Hybrid & 68.29 & 63.84 & 51.13 & 73.68 \\
Episodic-Outer & Hybrid & {\textbf{69.82}} & {\textbf{66.03}} & {\textbf{54.89}} & {\textbf{74.81}} \\
\bottomrule[1.0pt]
\end{tabular}
\caption{Performance over 133 test questions from three scenes. Latent uses no retrieved evidence. For Hybrid, both methods receive identical fixed retrieved evidence at test time.}
\label{tab:qa_only_outer_ablation}
\end{table}

As shown in Table~\ref{tab:qa_only_outer_ablation}, under Latent, the model answers without explicit evidence and must rely on its context-conditioned memory intuition. QA-only Outer is therefore well matched to this setting. It reinforces direct answering from the model's latent memory and encourages attention to the underlying logic between the context, the query, and the answer, which is why it outperforms Episodic-Outer by 3.50\% F1 and 3.01\% EM. Once evidence is introduced, however, the task imposes an additional requirement. The model must not only answer, but also correctly select, organize, and interpret precise memories. Episodic-Outer is better aligned with this requirement, since its outer-loop supervision already trains the model to assess and reconcile noisy episodic evidence into a more appropriate answer. This is reflected in the Hybrid setting, where Episodic-Outer surpasses QA-only Outer by 1.53\% F1 and 3.76\% EM. Since the Hybrid evidence is frozen and identical across the two runs, this gain cannot be attributed to changed retrieval.

Overall, Episodic-Outer achieves the best metrics under Hybrid, suggesting that effective memory use requires more than accurate evidence grounding. The model must also be able to interpret and integrate the retrieved evidence. These results further validate the rationality and effectiveness of our dual-memory framework.

\section{Adaptive Threshold Calibration}
\label{app:adaptive_threshold_calibration}

Our adaptive router combines two proxies: the normalized JSD divergence between the base and latent branches, and the latent branch's generation confidence. A query is routed to Hybrid when its divergence exceeds $\tau_d$; otherwise, it is routed to Latent only when its confidence exceeds $\tau_c$, and to Episodic in all remaining cases. 

Since different models and datasets exhibit their own preferred memory patterns or requirements, fixed thresholds can be suboptimal. We therefore calibrate one threshold pair $(\tau_d,\tau_c)$ shared by all tasks within each model--benchmark setting. Using the calibration queries and the already generated candidate answers, we select the pair that maximizes the mean Judge score of the routed answers. This experiment assesses whether the default thresholds $(0.1, 0.95)$ should be adjusted for a given setting.

\begin{wraptable}{r}{0.53\linewidth}
\vspace{-2em}
\centering
\small
\setlength{\tabcolsep}{5pt}
\begin{tabular}{l S[table-format=1.6] S[table-format=1.6]}
\toprule[1.0pt]
Setting & {\(\tau_d\)} & {\(\tau_c\)} \\
\midrule[0.5pt]
Mem-Gallery 7B & 0.082070 & 0.950000 \\
Mem-Gallery 3B & 0.003630 & 0.950000 \\
LoCoMo 7B & 0.401160 & 0.864015 \\
LoCoMo 3B & 0.015363 & 0.950000 \\
\bottomrule[1.0pt]
\end{tabular}
\caption{Shared calibrated thresholds for each model--benchmark setting. \(\tau_d\) is the normalized JSD threshold and \(\tau_c\) is the confidence threshold.}
\label{tab:adaptive_thresholds}
\end{wraptable}

As shown in Table~\ref{tab:adaptive_thresholds}, Mem-Gallery 7B yields $\tau_d=0.082070$ and $\tau_c=0.95$, close to the default values
$(0.1, 0.95)$. The calibration therefore preserves a similarly strict confidence requirement while making the divergence criterion only slightly more sensitive. Mem-Gallery 3B, by contrast, selects an
extremely low divergence threshold ($\tau_d=0.003630$) while retaining $\tau_c=0.95$, indicating that even minimal episodic--latent
disagreement favors Hybrid routing under a strict confidence requirement. The two LoCoMo settings differ more substantially. LoCoMo 7B selects a high divergence threshold ($\tau_d=0.401160$) together with a lower
confidence threshold ($\tau_c=0.864015$), indicating that Hybrid routing is reserved for stronger episodic--latent disagreement while more
responses can satisfy the confidence condition. LoCoMo 3B instead selects a very low divergence threshold ($\tau_d=0.015363$) while retaining $\tau_c=0.95$, so even small disagreement favors Hybrid routing without relaxing the confidence requirement.

Together, these results show that the appropriate routing thresholds are highly model- and benchmark-dependent, underscoring the need for an adaptive memory framework that can flexibly organize different memory sources across diverse tasks.

\end{document}